\documentclass[letterpaper, 10 pt, conference]{ieeeconf}  

\IEEEoverridecommandlockouts                              

\usepackage{graphics} 
\usepackage{epsfig} 
\usepackage{mathptmx} 
\usepackage{times} 
\usepackage{amsmath} 
\usepackage{amssymb}  
\usepackage{float}

\title{\LARGE \bf 
The NING Humanoid: The Concurrent Design and Development of a Dynamic and Agile Platform
}
\author{Yan Ning, Song Liu, Taiwen Yang, Liang Zheng, and Ling Shi 
\thanks{This work was supported by the Noetix Robotics company.}
\thanks{Y. Ning and L. Shi are with the Department of Electronic and Computer Engineering, the Hong Kong University of Science and Technology, Clear Water Bay, Hong Kong SAR
        {\tt\small (email: \{yningaa, eesling\}@connect.ust.hk).}}%
\thanks{S. Liu, T. Yang, and L. Zheng are with the Noetix Robotics company, Beijing, China
        {\tt\small (email: \{song.liu, taiwen.yang, zhengliang\}@noetixrobotics.com).}}%
}

\begin{document}
\maketitle
\thispagestyle{empty}
\pagestyle{empty}

\begin{abstract}

The recent surge of interest in agile humanoid robots achieving dynamic tasks like jumping and flipping necessitates the concurrent design of a robot platform that combines exceptional hardware performance with effective control algorithms. This paper introduces the NING Humanoid, an agile and robust platform aimed at achieving human-like athletic capabilities. The NING humanoid features high-torque actuators, a resilient mechanical co-design based on the Centroidal dynamics, and a whole-body model predictive control (WB-MPC) framework. It stands at 1.1 meters tall and weighs 20 kg with 18 degrees of freedom (DOFs). It demonstrates impressive abilities such as walking, push recovery, and stair climbing at a high control bandwidth. Our presentation will encompass a hardware co-design, the control framework, as well as simulation and real-time experiments.

\end{abstract}

\section{INTRODUCTION}

Recent years have witnessed a remarkable impact on research and industry fields due to the impressive capabilities demonstrated by humanoid robots like Atlas \cite{c1}, Digit \cite{c2}, and Optimus \cite{c3}. These advancements have promptly increased focus on the development of humanoid robots capable of executing operational tasks and dynamic motions such as running, jumping and flipping. However, the existing robot platforms still face limitations of flexibility and robustness for control. Thus, significant challenges remains in optimizing the concurrent design of hardware and control models to enhance robot dynamic performance.

This work presents the design and development of the NING humanoid, an agile and robust robot platform that achieves advanced dynamic performance. The NING humanoid combines a tightly integrated co-design based on the centroidal dynamic model \cite{c4,c5}, high-power-density actuators, and a WB-MPC-based controller. Notably, the NING humanoid demonstrates stable rough-terrain walking, push recovery, stair-climbing capabilities in experiments, and back-flipping in simulation. The subsequent sections present the hardware design, the control framework, and simulation and experiment results.

\begin{figure}
    \centering
    \includegraphics[scale = 0.17]{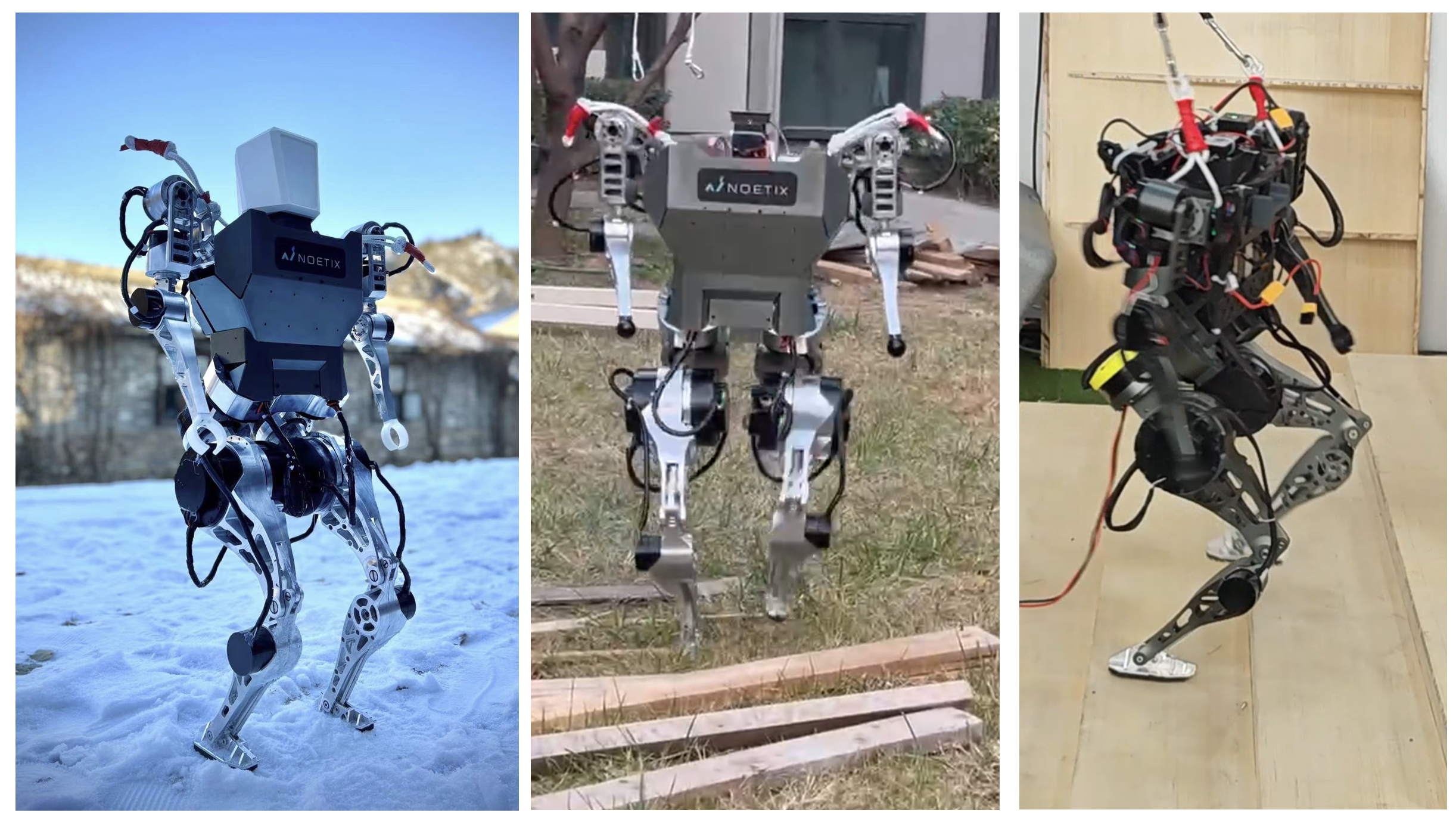}
    \caption{The NING humanoid platform at experimental scenarios.}
    \label{front_view}
\end{figure}

\section{RELATED WORKS}
Humanoids which perform extremely dynamic motions highly rely on the actuators. The Atlas performs advanced movements like backflips with hybrid-hydraulic actuators, while the Digit robot utilizes series-elastic actuators (SEAs) for precise torque control. Although these kinds of actuators demonstrate excellent performance, the research and maintenance cost is significant. Therefore, quasi-direct-drive (QDD) actuators are commonly applied in humanoids with low cost, high power density and smooth backdrivability at limb ends. Current QDD-based humanoids like the Unitree H1 \cite{c6} and the Artemis \cite{c7} from UCLA have already achieved large impact robustness and athletic capabilities. However, they may lose agility due to a large size, which may be dangerous for operators and large cost in maintenance.

Furthermore, the reduced system models, like the Spring-loaded-inverted-pendulum (SLIP) model \cite{c8} or the centroidal dynamic model \cite{c4,c9}, are commonly used to simplify the complex control calculation. Criteria shows that the inertia and mass of limbs and other parts significantly affect the approximation error in robot dynamics \cite{c10}. Thus, the design principle should concurrently consider the dynamic model and hardware layout to make the system more agile for control and improve the overall performance. In terms of transmission mechanism, the Digit robot uses multi-bar linkage \cite{c2} which is nonlinear and complicated for control, while the MIT humanoid employs synchronous belt between joints \cite{c11} which may lose some rigidity and affect the accuracy. Motivated by such limitations, we design and develop the NING humanoid achieving high-dynamic tasks with more agility, more robustness and low cost.

\section{HARDWARE DESIGN}
\begin{figure}
    \centering
    \includegraphics[scale = 0.18]{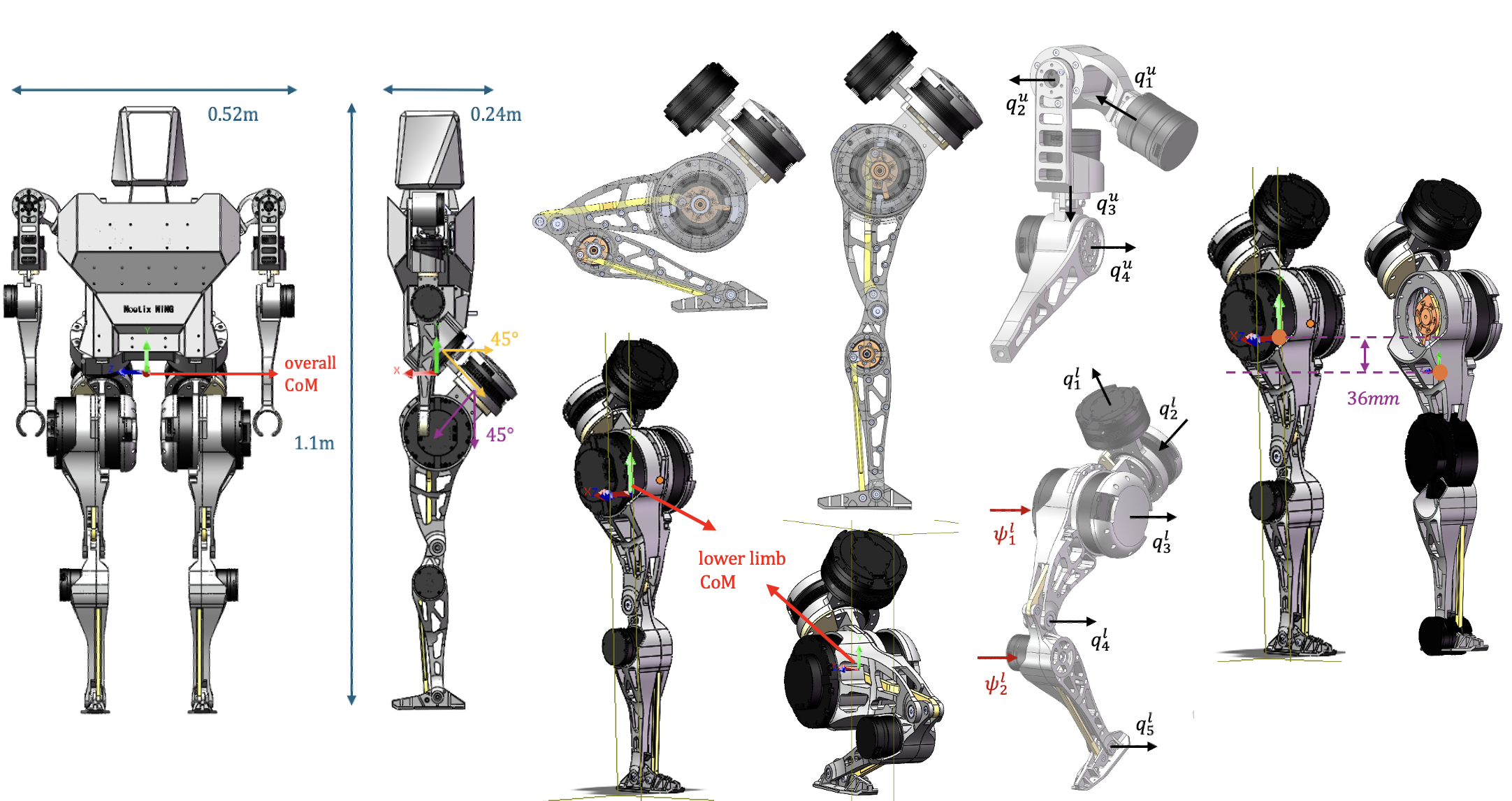}
    \caption{The design details of the NING humanoid, including robot size, transmission mechanism, axes and joints.}
    \label{details}
\end{figure}

The NING humanoid is designed to be robust for continuous impact and perform dynamic locomotion, which weighs near 20 kg and has 18 degree of freedom (DoFs) in total, 5 per leg and 4 per arm. An overview of size and joint arrangement is shown in Fig.~\ref{details}. The actuators operate at a high bandwidth as 1kHz through the Ethercat policy. An Intel NUC and an IMU are used for whole-body state estimation and real-time locomotion computation.

\subsection{Actuation and Mechanism}
There are three types of modular QDD actuators in the NING humanoid, corresponding to joint demands simulated based on tasks. Such actuators have low gear reduction ratios, high torque density and low cost. They largely improve the locomotion performance due to smooth back-drivability and accurate torque feedback at the joints. 

As the impact of contact between the foot and the ground is huge on every locomotion steps, high-strength mechanical parts and rigid transmission mechanism are required for hardware robustness. Therefore, most parts are manufactured as aluminum 7075 and the key stress components are machined with high stiffness steel. In order not to lose transmission linearity and mechanism rigidity, four bar linkage mechanism is utilized to pass the rotation and torque of actuators simultaneously to the knee and ankle joints shown in Fig.~\ref{details}. 

\subsection{Model-based Concurrent Design}
The approximation gap between the dynamic model and the real robot should be minimized to improve the locomotion agility and whole-body control accuracy. Based on the Centroidal Inertia Isotropy (CII) \cite{c10} and the centroidal momentum matrix \cite{c4}, the NING humanoid is co-designed to concentrate the heavy actuators close to the Center of Mass (CoM) of the whole body and reduce the weight of the limbs. The roll and yaw actuators at the hip are tilted to 45 degrees to give a more compact design of the crotch. It can be seen in Fig.~\ref{details} that the CoM of a single leg is quite near the CoM of the overall body, and it varies little at different postures. Moreover, the movement range of the upper and lower limbs are still wide to achieve diverse motions.

\section{CONTROL FRAMEWORK}
The WB-MPC control framework follows the work of the MIT Cheetah \cite{c12, c13} and Humanoid \cite{c14}. The MPC-based locomotion control uses the enhanced centroidal dynamics to predict the future states of the humanoid, while the WBC controller optimizes the current states and output joint torques and accelerations. A simple PD controller is applied for the motors to follow target indices of different actuators.

The locomotion MPC problem is formulated as a convex quadratic programming (QP) problem using the reduced centroidal dynamics to predict step placement of the robot, while the WBC problem is formulated as a convex QP using the full humanoid model. The WBC objective function is,
\begin{equation}\label{eq:1}
    \min_{\mathbf{\Ddot{q}}, \mathbf{\tau}, \mathbf{\mu}} \|\mathbf{A}_t \mathbf{\Ddot{q}} + \dot{\mathbf{A}}_t \mathbf{\dot{q}} - \dot{\mathbf{r}}_t\|^2 + \|\mathbf{u}^d - \mathbf{u}\|^2 + \|\mathbf{\tau}\|^2,
\end{equation}
subject to constraints including discrete linear dynamics, locomotion tasks regrading to the torse pose and joint angles, ground contact wrench cone, and actuator and power limits, where \(\mathbf{A}_t\) is the task Jacobian matrix; \(\dot{\mathbf{r}}_t\) is the commanded task dynamics; \(\mathbf{u}^d\) and \(\mathbf{u}\) are the desired and actual reaction wrench between the foot and the ground, and \(\mathbf{\tau}\) is the joint torques. By applying an efficient QP solver, qpOASES \cite{c15}, the WBC problem \eqref{eq:1} could be solved in real-time.

\section{SIMULATION \& EXPERIMENTS}
\begin{figure}
    \centering
    \includegraphics[scale = 0.2]{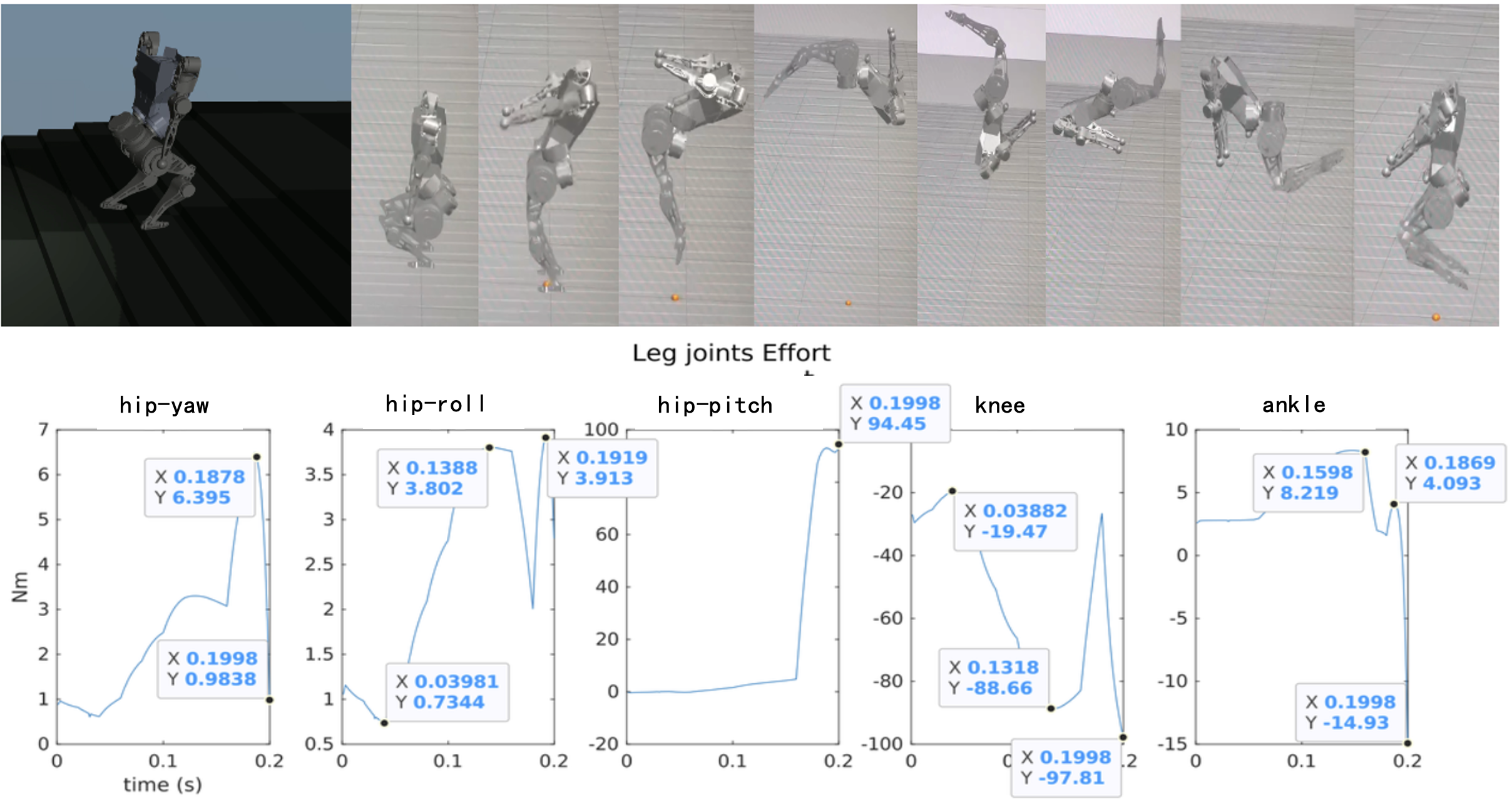}
    \caption{Illustrations of back-flipping in simulation and the results of different joint torques in each leg.}
    \label{simu_n_expe}
\end{figure}

The simulation is setup on the MuJoCo \cite{c16} platform, using a 3D urdf model with mass and inertia attached on the torso and limb axes. Different dynamic tasks like walking, stair-climbing and back-flipping are successfully simulated shown in Fig.~\ref{simu_n_expe}. The online controller is further fine-tuned for real-time implementation. 

In these experiments, the NING humanoid could switch states between stable standing and walking. It passes three dynamic locomotion scenarios shown as Fig.~\ref{front_view}: traversing rough grass and snow terrains, performing a push and slippery recovery, and climbing slopes at 10 degrees and low stairs. The gait pace could be self-adjusted based on the terrain reaction force feedback for the robot to resist unknown disturbances. 

\section{CONCLUSION}
In conclusion, to perform dynamic tasks, we design the NING humanoid as an agile and robust platform. Based on co-design principles, we integrate exceptional QDD actuators, concentrate inertia at joints based on robot dynamics, utilize linear four bar linkage as transmission to improve the control flexibility, and implement a WB-MPC controller. The simulation and experiments are conducted to show the performance of the NING humanoid. We will further augment perception-based state estimation and trajectory planning on the platform to achieve more advanced movements.




\section*{APPENDIX}

\subsection{Specification of the Actuators}
\begin{figure}[H]
    \centering
    \includegraphics[scale = 0.3]{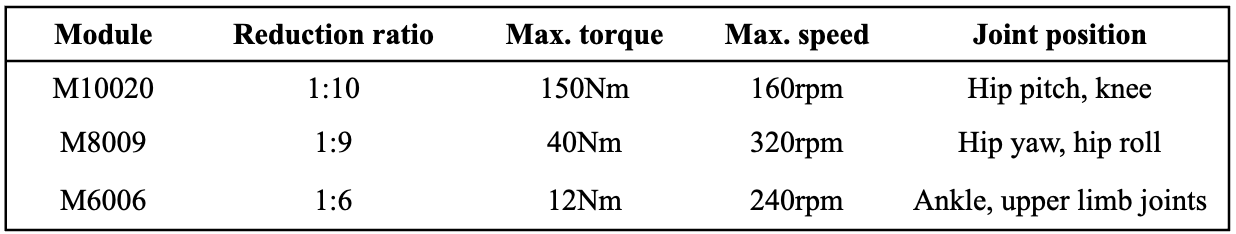}
    \caption{The specification each QDD actuator utilized in the NING humanoid}
    \label{motor_spec}
\end{figure}

\subsection{Comparison between Inertia of Different Designs}
\begin{figure}[H]
    \centering
    \includegraphics[scale = 0.23]{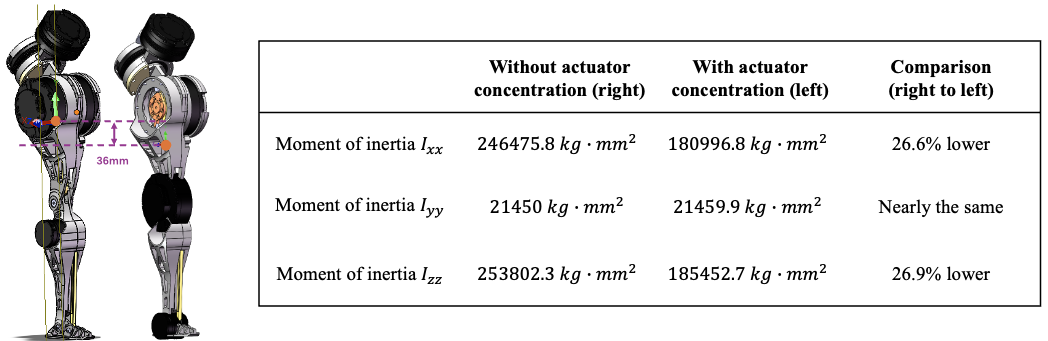}
    \caption{The comparison of CoM and momentum of inertia change based on different design principles.}
    \label{leg_inertia}
\end{figure}

\subsection{Illustrations of the Design and Experiments}
\begin{figure}[H]
    \centering
    \includegraphics[scale = 0.23]{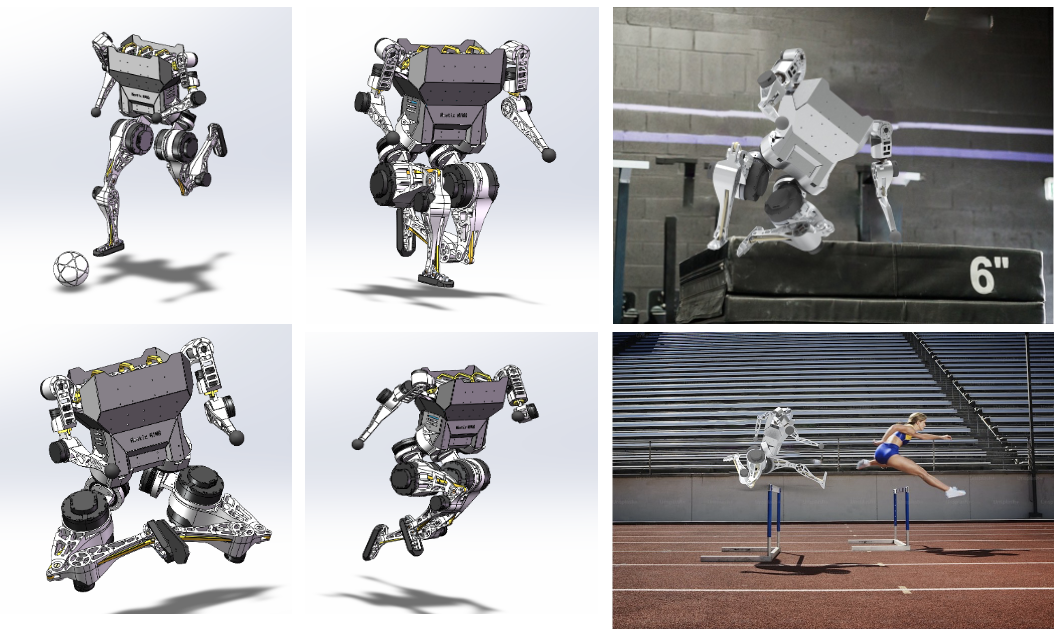}
    \caption{Illustrations of movements designed for the NING humanoid.}
    \label{some_designs}
\end{figure}

\begin{figure}[H]
    \centering
    \includegraphics[scale = 0.26]{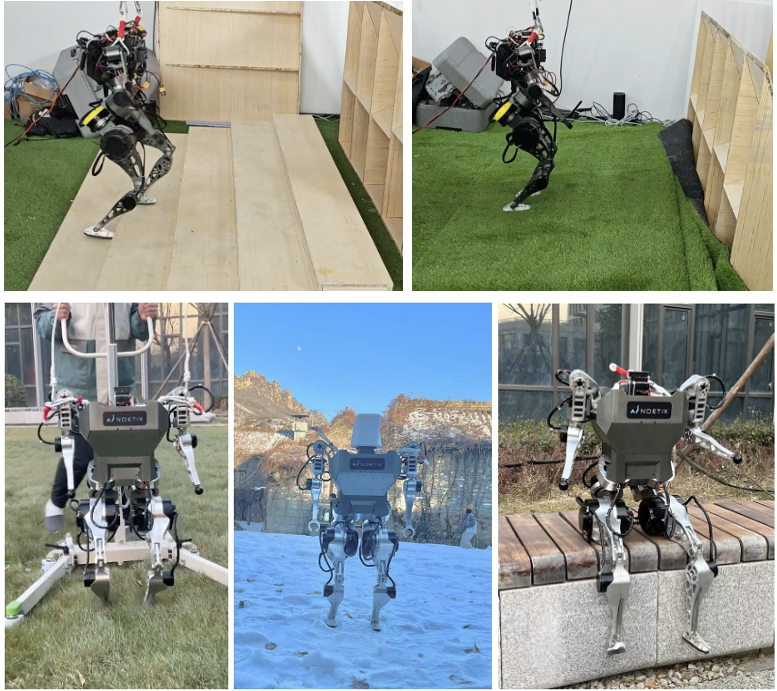}
    \caption{Illustrations of experiments conducted on the NING humanoid.}
    \label{some_expes}
\end{figure}

\subsection{The demonstration}
The demonstration video link is attached below, where the simulation and experiment are presented. All the rights are reserved by the Noetix company and the authors.

The video address: https://youtu.be/uWkuRK9mgyY.


\end{document}